%% file: main.tex
\documentclass[sigconf,screen,nonacm]{acmart}

\AtBeginDocument{%
  \providecommand\BibTeX{{%
    \normalfont B\kern-0.5em{\scshape i\kern-0.25em b}\kern-0.8em\TeX}}}

\input{headers/headers}

\begin{document}
\title{Towards Robust Graph Contrastive Learning}

\author{Nikola Jovanović}
\affiliation{%
  \institution{ETH Zurich}
  \city{Zurich}
  \country{Switzerland}
}
\email{njovanovic@ethz.ch}

\author{Zhao Meng}
\affiliation{%
  \institution{ETH Zurich}
  \city{Zurich}
  \country{Switzerland}
}
\email{zhmeng@ethz.ch}

\author{Lukas Faber}
\affiliation{%
  \institution{ETH Zurich}
  \city{Zurich}
  \country{Switzerland}
}
\email{lfaber@ethz.ch}

\author{Roger Wattenhofer}
\affiliation{%
  \institution{ETH Zurich}
  \city{Zurich}
  \country{Switzerland}
}
\email{wattenhofer@ethz.ch}

\begin{abstract}
  \input{abstract}
\end{abstract}

\maketitle

\input{intro}

\input{related-work}

\input{method}

\input{eval}

\input{conclusion}

\bibliographystyle{ACM-Reference-Format}
\bibliography{references}

\end{document}

%% file: headers/headers.tex
\usepackage[utf8]{inputenc} %
\usepackage{amsmath}
\usepackage{graphicx}
\usepackage{hyperref}

\usepackage{algorithm}
\usepackage[noend]{algpseudocode}

\usepackage[T1]{fontenc}

\newcommand{\ie}{i.e., }

\graphicspath{{figures/}}

\usepackage{csvsimple} %

\usepackage{booktabs}

\input{math_commands.tex}

\usepackage[capitalize]{cleveref}

\crefname{appendix}{Appendix}{Appendices}
\crefname{equation}{Equation}{Equations}
\crefname{section}{Section}{Sections}

\makeatletter
\newcommand\fs@booktabsruled{%
  \def\@fs@cfont{\bfseries\strut}\let\@fs@capt\floatc@ruled
  \def\@fs@pre{\hrule height\heavyrulewidth depth0pt \kern\belowrulesep}%
  \def\@fs@mid{\kern\aboverulesep\hrule height\lightrulewidth\kern\belowrulesep}%
  \def\@fs@post{\kern\aboverulesep\hrule height\heavyrulewidth\relax}%
  \let\@fs@iftopcapt\iftrue
}
\makeatother

\floatstyle{booktabsruled}
\restylefloat{algorithm}

\makeatletter
\def\algbackskip{\hskip-\ALG@thistlm}
\makeatother

\crefname{equation}{Eq.}{Eqs.}

%% file: math_commands.tex
\usepackage{amsmath,amsfonts,bm}
\usepackage{amsthm}

\def\1{\bm{1}}

\def\mA{{\bm{A}}}

\def\mX{{\bm{X}}}

\def\mZ{{\bm{Z}}}

\DeclareMathAlphabet{\mathsfit}{\encodingdefault}{\sfdefault}{m}{sl}
\SetMathAlphabet{\mathsfit}{bold}{\encodingdefault}{\sfdefault}{bx}{n}

\DeclareMathOperator*{\argmax}{arg\,max}

%% file: abstract.tex
We study the problem of adversarially robust self-supervised learning on graphs. In the contrastive learning framework, we introduce a new method that increases the adversarial robustness of the learned representations through i) adversarial transformations and ii) transformations that not only remove but also insert edges. We evaluate the learned representations in a preliminary set of experiments, obtaining promising results. We believe this work takes an important step towards incorporating robustness as a viable auxiliary task in graph contrastive learning.

%% file: intro.tex
\section{Introduction} \label{sec:intro}

Imagine we have unlimited labeled training data and a prediction task to solve. Due to the results from recent years, our first approach is to try a deep-learning-based model. In various domains, including computer vision, natural language processing,  or---more recently---graphs, deep learning models have proven to set the state of the art, given they receive sufficient data.

How should we proceed when there is not enough data? In this scenario, the picture becomes less clear. Generally, finding \emph{the data} is often not the main issue, but finding \emph{the labels} for this data is. Recent developments in deep learning attempt to tackle this problem by resorting to \emph{self-supervised learning}, where the labels are obtained by exploiting the internal structure of raw data. As one instance of this approach, \emph{contrastive learning} methods have recently achieved impressive results~\cite{chen2020simclr, khosla2020scl}. On a high level, contrastive learning attempts to learn the representations by applying transformations to the input without fundamentally changing it. The goal is to make the representations of a single input under different transformations similar, while the representations of different inputs should differ. The field of graphs lends itself particularly well to this setup. Notably, the Web enables us to mine massive graphs from, for example, the Web structure itself or social networks. On the other hand, labeling graph data is challenging as labels should reflect the complex network structure. \citet{zhu2020grace} show that we can use contrastive learning successfully in this setting.

Similarly to contrastive learning, the subject of \emph{adversarial attacks} also evolves around identity-preserving (\emph{imperceptible}) transformations. However, adversarial attacks aim to find imperceptible transformations that, despite looking innocent, cause a misprediction in the model. It was shown that even highly accurate neural networks are vulnerable to such attacks~\cite{intriguing-adv,adv,deepfool}, and thus unreliable, which is an especially important issue when they are used in safety-critical systems such as autonomous vehicles or face recognition systems. This lead to a great interest in building \emph{robust} models, \ie those less susceptible to adversarial attacks. In the contrastive learning setting, \citet{kim2020rocl} recently demonstrate that using \emph{adversarial transformations} allows us to learn robust representations. 
However, while the vulnerability to attacks is widely present in the graph domain \cite{dai2018rls2v,zugner2018nettack}, an investigation of adversarial transformations to learn robust representations is so far missing. In this paper, we explore this, and consider the use of adversarial transformations within the graph contrastive learning setting. We make the following contributions:

\input{figures/sketch}
\begin{enumerate}
    \item We propose \textbf{G}raph \textbf{Ro}bust \textbf{C}ontrastive Learning (\textbf{GROC}), a fully self-supervised graph algorithm aiming to achieve robustness to adversarial attacks. To the best of our knowledge, we are the first to integrate adversarial transformations into the graph contrastive learning framework.
    \item We conduct an evaluation of GROC on several popular transductive node classification datasets. The preliminary results show that GROC improves the robustness against adversarial attacks while maintaining a comparable performance on clean examples. 
    \item We outline possible future directions. We plan to extend our work to improve the efficiency of our method and extend our experiments to include a more comprehensive set of baselines and evaluation settings.
\end{enumerate}

%% file: figures/sketch.tex
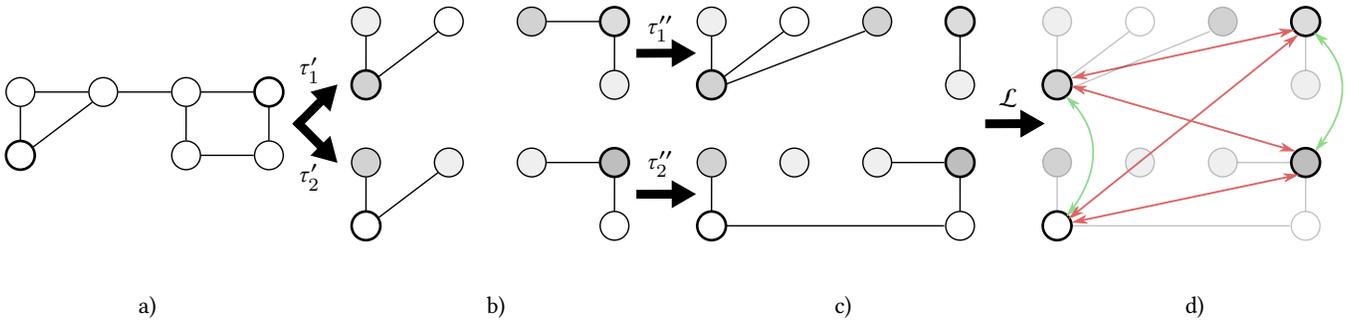
\begin{figure*}[t]
	\centering
    \def\svgwidth{\textwidth}
    \input{figures/groc_drawing}
	\caption{A high-level illustration of our method, GROC. (a) We start by choosing a set of anchor nodes from the graph. (b) Then, we apply two stochastic transformations $\tau_1'$ and $\tau_2'$ that randomly mask node features, to obtain two distinct views. The graph is reduced to the union of receptive fields of anchor nodes---here $1-$hop neighborhoods. (c) We apply adversarial transformations $\tau_1''$ and $\tau_2''$ to these two views, removing and inserting edges based on gradient signals. (d) Through contrastive learning, we force the embeddings of each anchor node in both views together (green arrows), and force the embeddings of different anchor nodes apart (red arrows).}
	\label{fig:sketch}
\end{figure*}
 

%% file: figures/groc_drawing.tex
\begingroup%
  \makeatletter%
  \providecommand\color[2][]{%
    \errmessage{(Inkscape) Color is used for the text in Inkscape, but the package 'color.sty' is not loaded}%
    \renewcommand\color[2][]{}%
  }%
  \providecommand\transparent[1]{%
    \errmessage{(Inkscape) Transparency is used (non-zero) for the text in Inkscape, but the package 'transparent.sty' is not loaded}%
    \renewcommand\transparent[1]{}%
  }%
  \providecommand\rotatebox[2]{#2}%
  \newcommand*\fsize{\dimexpr\f@size pt\relax}%
  \newcommand*\lineheight[1]{\fontsize{\fsize}{#1\fsize}\selectfont}%
  \ifx\svgwidth\undefined%
    \setlength{\unitlength}{541.79981292bp}%
    \ifx\svgscale\undefined%
      \relax%
    \else%
      \setlength{\unitlength}{\unitlength * \real{\svgscale}}%
    \fi%
  \else%
    \setlength{\unitlength}{\svgwidth}%
  \fi%
  \global\let\svgwidth\undefined%
  \global\let\svgscale\undefined%
  \makeatother%
  \begin{picture}(1,0.24150244)%
    \lineheight{1}%
    \setlength\tabcolsep{0pt}%
    \put(0,0){\includegraphics[width=\unitlength,page=1]{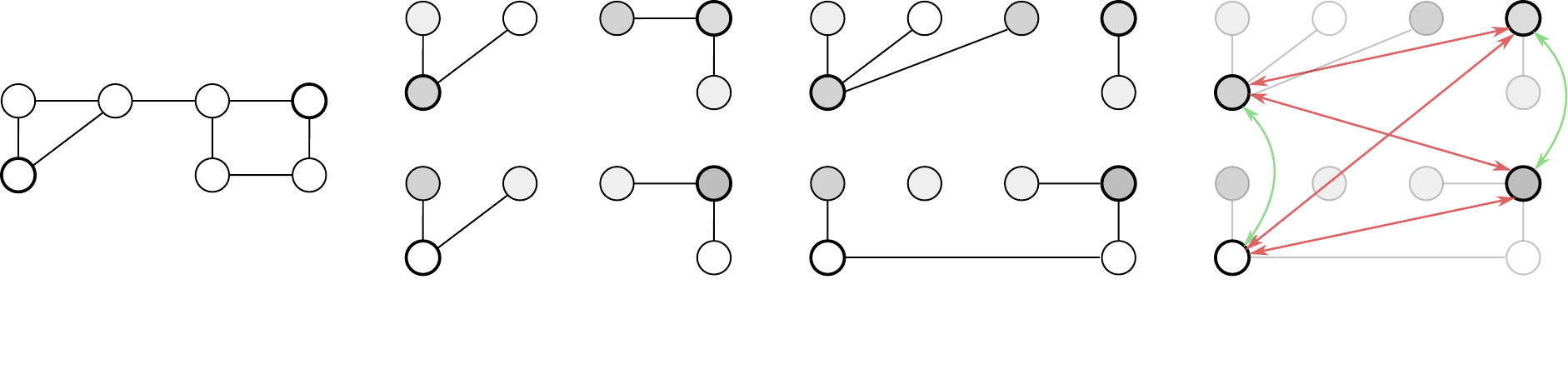}}%
    \put(0.10,0.01184204){\color[rgb]{0,0,0}\makebox(0,0)[lt]{\lineheight{1.25}\smash{\begin{tabular}[t]{l}a)\end{tabular}}}}%
    \put(0.36,0.01184204){\color[rgb]{0,0,0}\makebox(0,0)[lt]{\lineheight{1.25}\smash{\begin{tabular}[t]{l}b)\end{tabular}}}}%
    \put(0.62,0.01184204){\color[rgb]{0,0,0}\makebox(0,0)[lt]{\lineheight{1.25}\smash{\begin{tabular}[t]{l}c)\end{tabular}}}}%
    \put(0.882,0.01184204){\color[rgb]{0,0,0}\makebox(0,0)[lt]{\lineheight{1.25}\smash{\begin{tabular}[t]{l}d)\end{tabular}}}}%
    \put(0,0){\includegraphics[width=\unitlength,page=2]{groc.pdf}}%
    \put(0.22,0.19){\color[rgb]{0,0,0}\makebox(0,0)[lt]{\lineheight{1.25}\smash{\begin{tabular}[t]{l}$\tau_1'$\end{tabular}}}}%
    \put(0.22,0.11){\color[rgb]{0,0,0}\makebox(0,0)[lt]{\lineheight{1.25}\smash{\begin{tabular}[t]{l}$\tau_2'$\end{tabular}}}}%
    \put(0,0){\includegraphics[width=\unitlength,page=3]{groc.pdf}}%
    \put(0.48,0.221){\color[rgb]{0,0,0}\makebox(0,0)[lt]{\lineheight{1.25}\smash{\begin{tabular}[t]{l}$\tau_1''$\end{tabular}}}}%
    \put(0.48,0.12){\color[rgb]{0,0,0}\makebox(0,0)[lt]{\lineheight{1.25}\smash{\begin{tabular}[t]{l}$\tau_2''$\end{tabular}}}}%
    \put(0.7425,0.167){\color[rgb]{0,0,0}\makebox(0,0)[lt]{\lineheight{1.25}\smash{\begin{tabular}[t]{l}$\mathcal{L}$\end{tabular}}}}%
  \end{picture}%
\endgroup%

%% file: related-work.tex
\section{Related Work} \label{sec:relatedwork} 

\subsection{Graph Self-Supervised Learning}\label{sec:graph_ssl}

Early work on graph self-supervised learning focuses mostly on generative approaches. \citet{pretrain_gnn} propose to improve the performance of graph neural networks on downstream applications by using the tasks of link prediction, node ranking, and cluster recovery. \citet{hu2020gptgnn} propose the pretraining tasks of attribute and edge generation, improving model performance on downstream tasks, including node classification, link prediction, and community detection.

While the above approaches improve graph representations, recent breakthroughs in contrastive learning, mainly within computer vision~\cite{infomax, chen2020simclr}, motivate the study of similar algorithms in the graph domain~\cite{dgi,infograph,peng2020gmi,hassani2020multiview,qiu2020gcc,zhu2020grace,zhu2021gca,you2020graphcl}. In contrastive learning, the model is supposed to discriminate between positive (similar) and negative (dissimilar) pairs. In the formulation that we focus on,~\citet{zhu2020grace}, further improved in \citet{zhu2021gca}, create two views of the same graph by randomly removing edges and masking node features. In their work, positive pairs are the corresponding nodes in two views, while negative pairs are all other nodes in the same view (intra-view pairs) or the other view (inter-view pairs). \citet{you2020graphcl} use a similar procedure but focus on graph-level representations. Note that none of these works, as opposed to ours, have robustness as one of their goals during training.

Orthogonally to our contributions,~\citet{grill2020byol} recently suggest a novel self-supervised method that does not require negative pairs, improving the results of contrastive methods. This approach was subsequently adapted to the domain of graphs~\cite{che2020gbyol1,thakoor2021gbyol2}.
 
\subsection{Graph Adversarial Attacks \& Defenses}\label{sec:graph_adv}

The main insights regarding robustness, that even highly accurate networks are vulnerable to adversarial attacks, directly transfer to graphs. \citet{dai2018rls2v} propose RL-S2V, an attack which learns to create misclassifications through edge insertion and removal with reinforcement learning. Nettack~\cite{zugner2018nettack} crafts adversarial examples by perturbing the graph structure and altering the node features. \citet{zugner2019metaattack} propose Meta-Attack, employing meta-learning to produce adversarial examples.

In response to these attacks, researchers propose various \emph{graph purification} procedures~\cite{entezari2020gcnsvd,wu2019gcnjaccard}, as well as \emph{adversarial defenses}, training procedures designed to improve the robustness of neural networks, usually achieved by adversarially augmenting clean examples during training~\cite{pgd}. However, due to the discrete nature of edges and nodes in graphs, generating adversarial examples efficiently during training is hard~\cite{graph_adv_review, lat_gcn}.~\citet{dai2018rls2v} mitigate this problem by using examples with randomly dropped edges.~\citet{zhu2019rgcn} propose \emph{Robust GCN}, which absorbs the adversarial perturbations using Gaussian distributions as node representations in each layer of the network.~\citet{prognn} improve the adversarial robustness of graph neural networks by cleaning perturbed graphs through the intrinsic properties of real-world graphs such as low-rank adjacency matrices, sparse graphs, or homophily. Most recently,~\citet{zhang2020gnnguard} suggest a more general approach that is able to defend against attacks on heterophily graphs.

This goal of improving the robustness of networks extends to self-supervised contrastive learning methods as well. Recently, researchers argue that by using adversarial transformations during contrastive learning, a deep neural network model can achieve state-of-the-art robustness against image adversarial attacks~\cite{kim2020rocl, jiang2020acl, ho2020clae}. In the graph domain,~ \citet{you2020graphcl} include the evaluation of their contrastive learning method in the setting of adversarial attacks. However, they focus solely on graph classification, and more notably, they do not explicitly include adversarial robustness as a goal of their training.

%% file: method.tex
\section{Graph Robust Contrastive Learning} \label{sec:method} 

In this section, we introduce our method: \textbf{G}raph \textbf{Ro}bust \textbf{C}ontrastive Learning (\textbf{GROC}) (\cref{fig:sketch}). GROC builds on top of previous work in contrastive learning on graphs, aiming to improve graph neural networks' robustness against adversarial attacks.

\newcommand{\Tau}{\mathrm{T}}
\newcommand{\embed}[1]{\ensuremath{f_\theta(#1)}}

\input{figures/algo_groc}

\subsection{Background}
Consider a graph $\mathcal{G} = (V, \mX, \mA)$ with nodes $V=\{v_1, v_2, \ldots, v_n\}$, the node feature matrix $\mX \in \mathbb{R}^{n\times d}$, and the unweighted adjacency matrix $\mA \in \{0, 1\}^{n \times n}$. Our goal is to learn high-level representations (embeddings) of graph nodes $\mZ \in \mathbb{R}^{n \times d'}$ with $d' << d$. To this end, we learn a self-supervised encoder $f_\theta(\mathcal{G}) = \mZ$, where $f$ is a graph neural network parametrized by $\theta$. We denote the embedding of node $v$ as $\embed{v}$.

We train $f_\theta$ in the graph contrastive learning framework \cite{zhu2020grace,zhu2021gca,you2020graphcl} inspired by \citet{chen2020simclr}. The key idea is to treat $\mathcal{G}$ as merely \emph{one view} on the underlying input graph, not necessarily a unique one. We then define a family of identity-preserving transformations $T$, where two such transformations, $\tau_1, \tau_2 \in T$ map $\mathcal{G}$ to two new views of the same underlying graph, that is $\tau_1$ and $\tau_2$ do not change the fundamental structure of $\mathcal{G}$ and the node identities. Therefore, we expect the embeddings of the same node under $\tau_1$ and $\tau_2$ to be similar. At the same time, we expect the embeddings of different nodes to be dissimilar across and within two graph views. Let $Neg(v)=\{\tau_1(u) \mid u \in V\setminus \{v\}\} \cup \{\tau_2(u) \mid u \in V\setminus \{v\}\}$ be the embeddings of nodes other than $v$ in both graph views and $\sigma$ a similarity metric. We can obtain $\theta$ through the following optimization:
\begin{equation*}
	\argmax_\theta \mathbb{E}_{\tau_1, \tau_2 \sim \Tau} \left[\sum_{v \in V} \sigma(z_1, z_2) - \sum_{u \in Neg(v)} \sigma(z_1, \embed{u})\right],
\end{equation*}
where $z_1 \equiv \embed{\tau_1(v)}$ and $z_2 \equiv \embed{\tau_2(v)}$, and 

The above optimization is intractable to solve due to the massive search space of transformations $\Tau$ and a lack of an optimization algorithm. We follow the approach of \citet{zhu2020grace} to tackle this problem. We realize $\sigma$ as the cosine similarity between two embeddings after being fed through a $2-$layer MLP. Sampling two transformations $\tau_1, \tau_2$ from $T$, we can define a contrastive loss $\mathcal{L}(v,\tau_1,\tau_2)$ for each node as follows:
\begin{equation*}
\mathcal{L}(v, \tau_1, \tau_2) = - \log \frac{exp(\sigma(z_1, z_2)/t)}{exp(\sigma(z_1, z_2)/t) + \sum\limits_{u \in Neg(v)}exp(\sigma(z_1, \embed{u})/t)},
\end{equation*}
where $t$ is a temperature parameter. Finally, to derive a gradient-based update for $\theta$, we aim to minimize
\begin{equation} \label{eq:objective}
    \frac{1}{2n} \sum\limits_{v \in V} \left[ \mathcal{L}(v,\tau_1,\tau_2)+\mathcal{L}(v,\tau_2,\tau_1) \right].
\end{equation}
\citet{zhu2020grace} and \citet{zhu2021gca} follow this framework, considering the transformations that randomly remove a fraction of edges and randomly mask a fraction of node features with $0$. In GRACE~\cite{zhu2020grace}, the edges are removed uniformly. In GCA~\cite{zhu2021gca}, the authors investigate three variants where the edge removal probability is inversely proportional to the degree-based, eigenvector-based, or PageRank-based centrality scores of the edge.

\subsection{Motivation}
While the previously described contrastive learning methods obtain impressive results on a wide variety of tasks despite having no access to labels, their accuracy swiftly drops under adversarial attacks, as we later demonstrate in \cref{sec:evaluation}. 

Recall that the transformations used in contrastive learning aim to produce a view which is distinct from the input but is also \emph{imperceptible}, \ie the transformation should not fundamentally alter its identity, or in the case of graphs, the node identities. For most domains, various ways to define the notion of imperceptibility arise naturally, including $L_p$ norm perturbations and various image transformations. However, properly defining this notion for graphs is still an open challenge due to their discrete nature. Many methods resort to independent perturbations of features and edges, often simply performing random edge removal and random feature masking. In the context of adversarial defenses, we value transformations that increase the loss our optimization procedure attempts to minimize. We find the previously described choices lacking in this regard, which negatively impacts adversarial robustness. To partially alleviate this issue, we introduce several improvements to the choice of transformations $\tau_1, \tau_2$.

\subsection{Method}
We represent each $\tau_i \in T$ as a composition $\tau_i = \tau_i' \circ \tau_i''$. Namely, we generate two distinct views by first applying \emph{stochastic} transformations $\tau_1', \tau_2'$, followed by \emph{adversarial} transformations $\tau_1'', \tau_2''$. For $\tau_i'$ we simply employ random feature masking. For $\tau_i''$ we employ two types of edge-based transformations.

First, we perform edge removal as before. However, inspired by similar methods from other domains, we use the gradient information to make a more informed choice of edges to remove. Namely, we perform one preliminary forward-backward pass after applying $\tau_i'$ to obtain the gradients on edges. As we are minimizing \cref{eq:objective}, we remove a subset of edges with minimal gradient values. We discuss the choice of the number of edges to remove later.

Second, we introduce edge insertion, once again using the gradient information to choose the edges to insert. However, to obtain the gradients on the edges from our \emph{candidate set} for insertion $S^+$, we need to include those edges in the graph with a nonzero weight. Using all absent edges as the candidate set is impractical. To solve this, we tweak the training procedure by processing the nodes in randomized batches of size $b$, considering only the nodes in the current batch (\emph{anchors}) when constructing $Neg(v)$ and later \cref{eq:objective}. With this setup, we restrict $S^+$ to the set of edges $(u,v)$, where $v$ is an anchor node, and $u$ is within the $l$-hop neighborhood of some anchor $v' \neq v$, but not within the $l$-hop neighborhood of $v$. We temporarily insert all these edges into the graph with weights $1/|S^+|$, and after the preliminary pass, remove them, apart from a subset of edges with maximal gradient values. Note that $|S^+|$ is upper bounded by $nb$, which for small values of $b$ is a significant improvement over $\mathcal{O}(n^2)$, and additionally reduces the impact of the candidate edges on the result of the preliminary pass. Further, we hypothesize that there is an additional benefit to node batching, as this greatly reduces the number of negative examples in $Neg(v)$ for each $v$, focusing $v$ more on its representation in the other view. The GROC algorithm is illustrated in \cref{fig:sketch} and detailed in \cref{alg:groc}.

%% file: figures/algo_groc.tex
\begin{algorithm*}[t]
  \caption{Graph Robust Contrastive Learning (GROC)}
  \label{alg:groc}
  \begin{algorithmic}[1]
  \Statex \emph{Input}: $\mathcal{G}=(V,\mX \in \mathbb{R}^{n \times d},\mA \in \mathbb{R}^{n \times n})$ 
  \Statex \emph{Output}: Embedding matrix $\mZ$, for use in a downstream task
  \State Initialize the parameters $\theta$ of an $l$-layer GNN encoder $f$
  \State For each node $v$, precompute $V_l(v)$ and $E_l(v)$, the set of nodes (resp. edges) within $l$ hops of $v$ that affect its final embedding
  \For{$epoch=1,\ldots,n_{epochs}$}
  \State Randomly split $V$ in $n/b$ batches $B_i$ of size $b$
  \For{\textbf{each} batch $B_i$}
  \State Apply stochastic transformations $\tau_1',\tau_2'$ to $\mathcal{G}$ to obtain two views, masking features independently with probability $p_1$ (resp. $p_2$)
  \State $S^- \gets \bigcup_{v \in B_i} E_l(v)$ \Comment{Candidate set for edge removal}
  \State For each $v \in B_i$, $ \hat{V}_l(v) \gets \left( (\bigcup_{v' \in B_i} V_l(v')) \setminus V_l(v) \right)$, and $S^+ \gets \{(u, v) \mid v \in B_i, u \in \hat{V}_l(v) \}$ \Comment{Candidate set for edge insertion}
  \State Temporarily add $S^+$ to both views with weights $1 / |S^+|$, removing them after the next step
  \State For both views, apply $f$ and compute the contrastive loss as in \cref{eq:objective}, considering only the nodes in $B_i$
  \State Backpropagate the loss to obtain a gradient intensity value $g(e)$ at each edge $e$ 
  \State Further transform the views by applying the adversarial transformations $\tau_1''$ and $\tau_2''$, where each $\tau_i''$ removes $q^-_i \cdot |S^-|$ edges from \phantom{hackkk} $S^-$ with the minimal $g(e)$ and adds $q^+_i \cdot |S^+|$ edges from $S^+$ with the maximal $g(e)$
  \State For both views, apply $f$ and compute the contrastive loss as in \cref{eq:objective}, considering only nodes in $B_i$
  \State Update $\theta$ by applying gradient descent to minimize the contrastive loss
  \EndFor 
  \EndFor 
  \end{algorithmic}
\end{algorithm*}

%% file: eval.tex
\section{Experiments} \label{sec:evaluation}  
We provide a preliminary evaluation of GROC on a set of transductive node classification tasks. In this setting, given a partially labeled graph, the task consists of learning to fill in the missing labels. In our self-supervised setup, we first use the features of all nodes, but notably no labels, to learn high-level representations of each node. Then, we follow the linear evaluation protocol \cite{dgi}, and train a simple linear classifier on the produced embeddings, using the labels of training nodes for supervision.

We report standard classification accuracy and robust accuracy, so far assuming the threat model of Nettack \cite{zugner2018nettack}, a common targeted gray-box attack, here used in an evasion setting. We vary the perturbation budget from $1$ to $5$ and use a $2$-layer GCN \cite{kipf2017gcn} as the surrogate model. The robust accuracy is reported on a set of $10$ the most easily attacked nodes (the ones with the lowest surrogate margin), $10$ the least easily attacked nodes, and $20$ additional random nodes from the test set. In the future, we plan to investigate additional attack methods, such as RL-S2V~\cite{dai2018rls2v} or Meta-Attack~\cite{zugner2019metaattack}.

For our implementation we use PyTorch \cite{pytorch}, heavily relying on the PyTorchGeometric \cite{geometric} library. For Nettack, we use the reference implementation from the DeepRobust~\cite{li2020deeprobust} library. We perform all experiments on a single GPU.

\subsection{Networks and Datasets}
As our encoder $f_\theta$, we use a $2$-layer GCN with layer sizes respectively $2n_h$ and $n_h$, and activation $act$, with the concrete values shown in \cref{table:hyperparams}. We evaluate on the following five datasets:
\begin{itemize}
    \item The standard citation network benchmarks \emph{Cora}, \emph{Citeseer} and \emph{Pubmed} \cite{datasets}, where nodes and edges represent the documents and the citations between them. We follow the dataset splits of \citet{kipf2017gcn} based on the setup of \citet{corasplit}, using $20$ nodes per class for training, $500$ nodes for the validation, and $1000$ nodes for the test set.
    \item \emph{AmazonPhoto} \cite{amazondatasets}, a segment of the Amazon co-purchase graph. Nodes represent products, and edges imply that two products are frequently purchased together. We randomly construct the training and validation sets with $10\%$ of the nodes each; the remaining nodes constitute the test set.
    \item \emph{WikiCS} \cite{mernyei2020wikics}, a dataset of computer science Wikipedia articles with edges based on hyperlinks between them. We use all $20$ dataset splits provided in the original paper and report the average results.
\end{itemize}
To meet our evaluation's assumptions, we preprocess each dataset if needed to ensure that the features are binary (thresholding at 0) and the graph is undirected with no multiple edges.

\subsection{Investigated Methods}
We compare GROC against two other self-supervised methods, GRACE~\cite{zhu2020grace} and GCA~\cite{zhu2021gca}. We show comparisons with the degree-based variant of GCA (GCA-DE), as we did not observe significant differences compared to other variants. Furthermore, we evaluate GRACE-ADV, an extension of GRACE that, instead of removing edges randomly, uses gradient signals as in GROC. Finally, we include a fully supervised baseline GCN, where we train the entire network with the supervision from node labels.
\input{figures/table_results}
\input{figures/table_results2}
\subsection{Hyperparameters} \label{ssec:hyperparams}
For GCN, we match the setup of \citet{kipf2017gcn}, using a $2$-layer GCN with the hidden layer size of $16$. For all self-supervised baselines we follow the hyperparameter choices from GRACE and GCA to choose the GCN parameters ($n_h$ and $act$), the training parameters ($n_{epochs}$, the initial learning rate for the Adam optimizer $\eta$, and the L2 penalty parameter $\lambda$), the contrastive loss temperature $\tau$, as well as the feature masking rates $p_1$ and $p_2$ and the edge removal rates $q^-_1$ and $q^-_2$. Notably, for GRACE-ADV we reduce $q^-_1$ and $q^-_2$ on two datasets as necessary to obtain convergence.

For GROC, we use the same set of hyperparameters, again reducing $q^-_1$ and $q^-_2$ as in GRACE-ADV and additionally using significantly fewer epochs for training as it converges early. Additionally, GROC introduces three more hyperparameters: the edge insertion rates $q^+_1$ and $q^+_2$ and $b$, the size of node batches. We experimentally find that a set of $10$ anchor nodes per batch works well across all datasets. We tune the edge insertion rates separately for each dataset. In \cref{table:hyperparams} we show all hyperparameter choices.
\subsection{Results}

The results of our experiments are shown in \cref{table:results,table:results2}. As a sanity check, we see that all self-supervised methods achieve standard accuracy comparable with the supervised GCN. As a second sanity check, we see that the accuracy rapidly drops for all baseline methods as we increase the perturbation budget of Nettack from $1$ to $5$. This confirms that they are indeed susceptible to adversarial attacks and motivates our focus on building a more robust self-supervised method.

Notice that GRACE-ADV often shows an improvement in adversarial robustness over the vanilla version of GRACE, demonstrating the efficacy of introducing adversarial transformations into the training framework, \ie removing those edges that hurt the contrastive loss the most, instead of randomly. 

Finally, we observe that the combination of edge insertion and adversarial transformations leads to the most robust method overall, GROC, which consistently boosts the robustness of the baseline methods. Note that graph attacks often rely heavily on edge insertion\footnote{We further empirically confirm this by directly examining the attacks performed by Nettack.}, which explains its importance as part of the transformation in the contrastive learning setting. GROC is the only method that learns to incorporate adversarial edge insertion into its representation, thus allowing for higher adversarial robustness.

\input{figures/table_hyperparams}

\subsection{Limitations}

While GROC succeeds in improving the adversarial robustness of graph contrastive learning methods, we point out that more work is needed, as the robustness of GROC currently comes at a price. While, due to batching, one epoch carries more information (and we can thus converge in fewer epochs), the complete training procedure of GROC takes an order of magnitude longer in total time. Therefore, we identify the work on optimizing and improving the method's performance as an important direction for future work. 

Additionally, to make the evaluation thorough and gain more complete insights into the proposed method's effects, we plan to include more datasets, inductive classification settings, and evaluate the adversarial robustness under a larger set of adversarial attacks. We further plan to include supervised adversarial defenses~\cite{lat_gcn, graphat} in our evaluation.

%% file: figures/table_results.tex
\begin{table*}[t]\centering
    \begin{minipage}{\textwidth}\caption{The results on Cora, Citeseer and Pubmed datasets. $Acc$ denotes standard accuracy, $N$ represents the robust accuracy under Nettack with perturbation budget $N$.}  \label{table:results} \end{minipage}
    
    \newcommand{\sixcol}[1]{\multicolumn{6}{c}{#1}}
    \renewcommand{\arraystretch}{1.2}
  
    \small
  
    \begin{tabular}{@{}c cccccc c cccccc c cccccc@{}} \toprule
    & \sixcol{Cora} & \phantom{a} & \sixcol{Citeseer} & \phantom{a} & \sixcol{Pubmed} \\ \cmidrule{2-7} \cmidrule{9-14} \cmidrule{16-21}
    Method & Acc & 1 & 2 & 3 & 4 & 5 && Acc & 1 & 2 & 3 & 4 & 5 && Acc & 1 & 2 & 3 & 4 & 5 \\ \midrule
    \begin{filecontents*}{results/results.csv}
Method,CoraAcc,CoraA,CoraB,CoraC,CoraD,CoraE,CiteAcc,CiteA,CiteB,CiteC,CiteD,CiteE,PubmedAcc,PubmedA,PubmedB,PubmedC,PubmedD,PubmedE
GCN,81.5,66.0,47.5,39.5,32.5,26.0,68.9,57.5,44.0,39.5,29.5,22.5,79.4,50.0,40.0,33.5,30.5,27.0
GRACE,81.6,63.0,47.5,40.0,30.0,23.5,69.8,61.5,49.5,39.0,33.0,{\bf 30.5},82.9,49.0,38.5,33.0,30.5,29.0
GCA-DE,{\bf 82.4},63.5,47.5,39.0,30.0,23.5,68.8,61.5,49.5,39.0,33.0,29.0,83.5,50.5,36.0,31.0,29.5,27.0
GRACE-ADV,{\bf 82.4},65.5,48.0,41.0,32.0,27.5,{\bf 70.7},62.5,50.5,38.5,33.0,{\bf 30.5},{\bf 84.0},48.5,40.5,36.0,33.5,32.0
GROC,77.0,{\bf 69.0 },{\bf 58.5 },{\bf 47.5},{\bf 42.5 },{\bf 37.5 },67.6,{\bf 67.5},{\bf 58.0},{\bf 45.5},{\bf 35.5},30.0,83.3,{\bf 53.5},{\bf 43.5},{\bf 38.5},{\bf 35.0},{\bf 32.5}
\end{filecontents*}
\csvreader[%
    head to column names, 
    late after line=\ifcsvstrcmp{\Method}{GRACE}{\\\midrule}{\\},
    ]{results/results.csv}{}%
    {\Method & \CoraAcc & \CoraA & \CoraB & \CoraC & \CoraD & \CoraE && \CiteAcc & \CiteA & \CiteB & \CiteC & \CiteD & \CiteE && \PubmedAcc & \PubmedA & \PubmedB & \PubmedC & \PubmedD & \PubmedE}%
    \bottomrule
    \end{tabular}
\end{table*}

%% file: figures/table_results2.tex
\begin{table*}[t]\centering
    \begin{minipage}{\textwidth}\caption{The results on AmazonPhoto and WikiCS datasets. $Acc$ denotes standard accuracy, $N$ represents the robust accuracy under Nettack with perturbation budget $N$.} \label{table:results2} \end{minipage} 
    
    \newcommand{\sixcol}[1]{\multicolumn{6}{c}{#1}}
    \renewcommand{\arraystretch}{1.2}
  
    \small
  
    \begin{tabular}{@{}c cccccc c cccccc@{}} \toprule
    & \sixcol{AmazonPhoto} & \phantom{abc} & \sixcol{WikiCS} \\ \cmidrule{2-7} \cmidrule{9-14}
    Method & Acc & 1 & 2 & 3 & 4 & 5 && Acc & 1 & 2 & 3 & 4 & 5 \\ \midrule
    \begin{filecontents*}{results/results2.csv}
Method,AphotoAcc,AphotoA,AphotoB,AphotoC,AphotoD,AphotoE,WikiAcc,WikiA,WikiB,WikiC,WikiD,WikiE
GCN,91.5,66.0,58.5,49.5,46.0,41.0,77.4,62.1,53.0,47.3,43.8,42.0
GRACE,{\bf 92.3},71.5,62.5,56.5,53.0,49.0,79.0,63.8,55.6,50.2,45.9,43.6
GCA-DE,91.7,70.0,63.0,57.0,53.0,{\bf 50.5},79.1,64.8,55.8,50.2,46.0,43.5
GRACE-ADV,91.0,72.0,63.5,56.5,{\bf 54.5},50.0,{\bf 79.2},{\bf 65.2},55.4,50.0,46.6,44.3
GROC,91.3,{\bf 72.5},{\bf 64.0},{\bf 57.5},53.5,{\bf 50.5},77.5,64.5,{\bf 57.6},{\bf 54.0},{\bf 50.0},{\bf 47.1}
\end{filecontents*}
\csvreader[%
    head to column names, 
    late after line=\ifcsvstrcmp{\Method}{GRACE}{\\\midrule}{\\},
    ]{results/results2.csv}{}%
    {\Method & \AphotoAcc & \AphotoA & \AphotoB & \AphotoC & \AphotoD & \AphotoE && \WikiAcc & \WikiA & \WikiB & \WikiC & \WikiD & \WikiE}%
    \bottomrule
    \end{tabular}
\end{table*}

%% file: figures/table_hyperparams.tex
\begin{table*}[t]\centering
    \begin{minipage}{\textwidth}\caption{Hyperparameter settings for all datasets. Values apart from $q^+_1$, $q^+_2$, and $b$ are taken from GRACE/GCA. The values of $q^-_1$ and $q^-_2$ marked with an asterisk are reduced to $0.01$ in GROC and GRACE-ADV. $n_{epochs}$ shows the reduced value used for GROC in parentheses.} \label{table:hyperparams} \end{minipage} 
    
    \newcommand{\sixcol}[1]{\multicolumn{6}{c}{#1}}
    \renewcommand{\arraystretch}{1.2}
  
    \small
  
    \begin{tabular}{@{}cccccccccccccc@{}} \toprule
    Dataset & $n_{h}$ & $act$ & $n_{epochs}$ & $\eta$ & $\lambda$ & $\tau$ & $p_1$ & $p_2$ & $q^-_1$ & $q^-_2$ & $q^+_1$ & $q^+_2$ & $b$ \\ \midrule
    \begin{filecontents*}{results/hyperparams.csv}
Dataset,NH,Act,Epochs,LR,WD,Tau,FtOne,FtTwo,DropOne,DropTwo,AddOne,AddTwo,B
Cora,128,relu,200 (200),0.0005,1e-5,0.4,0.3,0.4,0.2,0.4,0.004,0.004,10
Citeseer,256,prelu,200 (200),0.001,1e-5,0.9,0.3,0.2,0.2,0,0.04,0.04,10
Pubmed,256,relu,1500 (300),0.001,1e-5,0.7,0,0.2,0.4,0.1,0.004,0.004,10
AmazonPhoto,256,prelu,1500 (100),0.01,1e-5,0.7,0.1,0.2,0.4*,0.1*,0.0004,0.0004,10
WikiCS,256,prelu,3000 (200),0.01,1e-5,0.7,0.2,0.1,0.2*,0.3*,0.004,0.004,10
\end{filecontents*}
\csvreader[%
    head to column names, 
    late after line=\\,
    ]{results/hyperparams.csv}{}%
    {\Dataset & \NH & \Act & \Epochs & \LR & \WD & \Tau & \FtOne & \FtTwo & \DropOne & \DropTwo & \AddOne & \AddTwo & \B} %
    \bottomrule
    \end{tabular}
\end{table*}

%% file: conclusion.tex
\section{Conclusion} \label{sec:conclusion}

In this work, we focused on the issue of adversarial robustness of self-supervised learning methods on graphs. We suspected, and later experimentally confirmed, that the previously introduced contrastive learning methods are vulnerable to adversarial attacks. As a first step towards achieving robustness in this setting, we introduced a novel method, GROC, that enhances the generation of graph views by introducing adversarial transformations and edge insertion. We confirmed that this approach can improve the adversarial robustness of the produced representations through a preliminary set of experiments. We hope that this work will ultimately lead to more successful and more robust contrastive learning algorithms on graphs.
\bigskip
\balance %